\documentclass[letterpaper, preprint, paper, 11pt]{AAS}

\usepackage{ifluatex}
\ifluatex
  \usepackage{pdftexcmds}
  \makeatletter
  \let\pdfstrcmp\pdf@strcmp
  \let\pdffilemoddate\pdf@filemoddate
  \makeatother
\fi
\usepackage{svg}
\usepackage[utf8]{inputenc}
\usepackage{textcomp}
\usepackage{multicol}
\usepackage{lscape}
\usepackage{algorithm}
\usepackage{algpseudocode}
\usepackage{svg}
\usepackage{enumitem}

\usepackage{xcolor}
\usepackage{soul}

\usepackage{epstopdf} %converting to PDF
\usepackage{algorithm}
\usepackage{algpseudocode}
\usepackage{graphicx}
\usepackage{amsmath}
\usepackage[version=4]{mhchem}
\usepackage{siunitx}
\usepackage{longtable,tabularx}
\usepackage{xcolor} % font colors
\usepackage{wrapfig} % wrap figures
\usepackage{subcaption} % for subfigures
\usepackage{array}
\newcolumntype{L}[1]{>{\raggedright\let\newline\\\arraybackslash\hspace{0pt}}m{#1}}
\newcolumntype{C}[1]{>{\centering\let\newline\\\arraybackslash\hspace{0pt}}m{#1}}
\newcolumntype{R}[1]{>{\raggedleft\let\newline\\\arraybackslash\hspace{0pt}}m{#1}}

\usepackage{hyperref}
\hypersetup{
    colorlinks=true,
    linkcolor=black,
    anchorcolor=black,
    citecolor=black,
    filecolor=black,
    menucolor=black,
    runcolor=black,
    urlcolor=blue
}

\usepackage{cleveref}

\usepackage{nomencl}
\usepackage{amsfonts}
\makenomenclature

\title{Deep Reinforcement Learning for Autonomous Spacecraft Inspection using Illumination\thanks{\NoCaseChange{Approved for Public Release, Case Number AFRL-2023-1932. The views expressed are those of the authors and do not reflect the official guidance or position of the United States Government, the Department of Defense or of the United States Air Force.}}}

\author{David van Wijk\thanks{Graduate Research Fellow, Aerospace Engineering, Texas A\&M University, Land, Air, and Space Robotics (LASR) Laboratory, 1188 Nuclear Science Rd, College Station, TX, 77845.}, Kyle Dunlap\thanks{AI Scientist, RDT\&E Division, Parallax Advanced Research, 4035 Colonel Glenn Hwy, Beavercreek, OH, 45431.}, Manoranjan Majji\thanks{Associate Professor, Director, Land, Air, and Space Robotics (LASR) Laboratory, Aerospace Engineering, Texas A\&M University, 1188 Nuclear Science Rd, College Station, TX, 77845.},\ and Kerianne L. Hobbs\thanks{Safe Autonomy Lead, Autonomy Capability Team 3, Air Force Research Laboratory, 2241 Avionics Circle, Wright-Patterson AFB, OH, 45433.}}

\PaperNumber{23-342}

\begin{document}

\maketitle 

\begin{abstract}
\noindent This paper investigates the problem of on-orbit spacecraft inspection using a single free-flying deputy spacecraft, equipped with an optical sensor, whose controller is a neural network control system trained with Reinforcement Learning (RL). This work considers the illumination of the inspected spacecraft (chief) by the Sun in order to incentivize acquisition of well-illuminated optical data. The agent's performance is evaluated through statistically efficient metrics. Results demonstrate that the RL agent is able to inspect all points on the chief successfully, while maximizing illumination on inspected points in a simulated environment, using only low-level actions. Due to the stochastic nature of RL, 10 policies were trained using 10 random seeds to obtain a more holistic measure of agent performance. Over these 10 seeds, the interquartile mean (IQM) percentage of inspected points for the finalized model was 98.82\%.
\end{abstract}

\section{Introduction}

Autonomous spacecraft inspection is foundational to sustained, complex spacecraft operations and uninterrupted delivery of space-based services. Inspection may enable investigation and characterization of space debris, or be the first step prior to approaching a prematurely defunct satellite to repair or refuel it.   Additionally, it may be mission critical to obtain accurate information for characterizing vehicle condition of a cooperative spacecraft, such as in complex in-space assembly missions.

The common thread among all the potential applications is the need to gather information about the resident space object, which can be achieved by inspecting the entire surface of the object. As such, this paper addresses the problem of inspecting the entire surface of a chief spacecraft, using a simulated imaging sensor on a free-flying deputy spacecraft. In particular, this research considers illumination requirements for optical sensors.  

A machine learning-based approach, namely Reinforcement Learning (RL), is used to generate a solution to this problem. Learning-based algorithms for control can be preferred over traditional optimization-based control algorithms because they allow the bulk of the computation to be done offline and are generally light and computationally-inexpensive when deployed. For example, a large set of optimized control decisions used by Airborne Collision Avoidance System X was compressed with a Neural Network (NN), reducing the storage space by a factor of 1000 \cite{ACAS2016}.

In recent years there have been many successful attempts to use deep learning techniques for spacecraft control applications in simulated environments. Dunlap et al. demonstrated the effectiveness of a RL controller for spacecraft docking in tandem with run time assurance (RTA) methods to assure safety \cite{Dunlap2023}. Gaudet et al. proposed an adaptive guidance system using reinforcement meta-learning for various applications including a Mars landing with random engine failure \cite{Gaudet2019}. The authors demonstrate the effectiveness of their solution by outperforming a traditional energy-optimal closed-loop guidance algorithm developed by Battin \cite{Battin1987AnIT}. Campbell et al. developed a deep learning structure using Convolutional Neural Networks (CNNs) to return the position of an observer based on a digital terrain map, meaning that the pre-trained network can be used for fast and efficient navigation based on image data \cite{Campbell2017}. Similarly, Furfaro et al. use a set of CNNs and Recurrent Neural Networks (RNNs) to relate a sequence of images taken during a landing mission, and the appropriate thrust actions \cite{Furfaro2018}.  

Similarly, previous work has been done to solve the inspection problem using both learning-based and traditional methods. In a recent study by Lei et al., the authors use deep RL to solve the inspection problem using multiple 3-Degree-of-Freedom (DOF) agents and hierarchical RL \cite{LeiGNC22}. They split the inspection task into two sub-problems: 1) a guidance problem, where the agents are assigned waypoints that will result in optimal coverage, and 2) a navigation problem, in which the agents perform the necessary thrusting maneuvers to visit the points generated in 1). The solutions to the two separate problems are then joined and deployed in unison. Building on this work, Aurand et al. developed a solution for the multi-agent inspection problem of a tumbling spacecraft, but approached this problem by considering collection of range data instead of visiting specific waypoints \cite{AurandGNC23}. In a very similar application to this paper, Brandonisio et al. used an RL based approach to map an uncooperative space object using a free-flying 3 DOF spacecraft \cite{Brandonisio2021}. While the authors consider the role of the Sun in generating useful image data, they do so using fixed logic based on incidence angles, rather than an explicit technique such as the ray-tracing technique proposed here.

However, as far as the authors know, there has not been an attempt to solve the inspection problem by incorporating illumination on the spacecraft of interest using ray tracing derived techniques. As such, the central contributions of this work are two-fold. Firstly, an integrated physics based model for realistic illumination is built.
Secondly, the effectiveness of using a RL-trained controller to inspect a chief with the realistic illumination model is demonstrated.

The remainder of this manuscript is organized as follows. First, the background theory necessary for a high-level RL understanding will be introduced. Following this will be a problem definition, with the governing dynamics and assumptions clearly defined. Next, the mechanics of the underlying illumination model will be highlighted, followed by a detailed description of the simulation setup and RL specific parameters such as the actions, rewards and observations available to the agent. Finally, the results are presented demonstrating the effectiveness of the solution accompanied by an analysis and future considerations for the problem.

\section{Reinforcement Learning}
This section provides a brief introduction to the Reinforcement Learning (RL) problem which is used to solve the illumination-aware inspection problem. 

\begin{figure}[H]
    \centering
    % \centerline{\includesvg[inkscapelatex=false,width=0.6\columnwidth]{figs/general_RL.svg}}
    \includegraphics[angle=0,origin=c,width=0.6\textwidth]{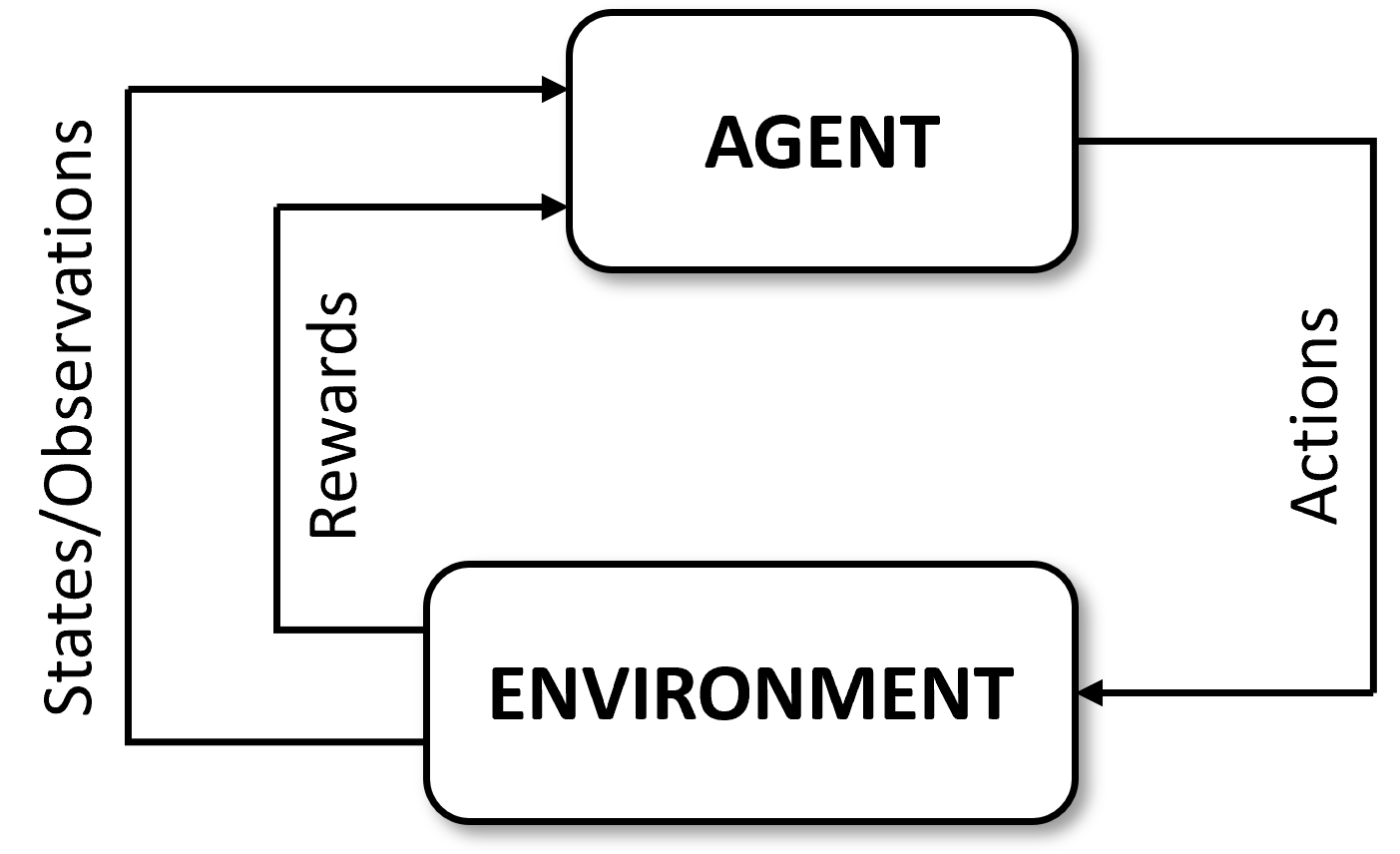}
    \caption{Diagram of general reinforcement learning problem.}
    \label{RL_diagram}
\end{figure}

RL is built on a trial and error paradigm, where the agent increases its performance based on rewarded behavior collected while interacting with the environment as shown in Fig.~\ref{RL_diagram}. Deep Reinforcement Learning (DRL) is a newer branch of RL in which a neural network is used to approximate the behavior function,
i.e. policy $\pi$, learned by the agent. The policy maps the observation vector $\boldsymbol{o}$ of a partially-observable environment to the action vector $\boldsymbol{a}$ of the agent. In a fully-observable environment, the observation is the full state information given by state vector $\boldsymbol{s}$. The agent is evaluated using the scalar reward signal $r$, which, along with the states, is an output of the environment. The DRL algorithm learns the optimal policy that maximizes the cumulative reward.

The specific RL algorithm used in this work is Proximal Policy Optimization (PPO) \cite{PPO}. PPO is a policy gradient method, that uses two neural networks, an \textit{actor} and a \textit{critic}. The critic learns the value function, which estimates the expected cumulative reward that will be accrued following the current policy from the current state. The actor is updated by increasing the probability of choosing a particular action if the received reward is higher than what was predicted by the critic's value function. Conversely, it will decrease the probability of selecting an action if the received reward is lower than what was expected. The actor is the NN controller that is eventually deployed once the training is complete. Unlike other policy gradient methods, PPO limits large updates from previous ``trusted" policies in order to avoid rapid decreases in performance which are known as cases of ``catastrophic forgetting". An earlier DRL algorithm, \textit{Trust Region Policy Optimization} (TRPO) \cite{schulman2015trust} developed this concept, but solves an optimization problem to obtain the largest change to the policy that still remains within a ``trustworthy" region. PPO on the other hand uses a clipping function with a hard limit to simplify this process, speeding up computation. PPO was selected for this application because of its history producing high-performing solutions \cite{Dunlap2023,hamilton2022ablation,andrychowicz2020matters} and ease of use. 

\section{Problem Formulation}

The spacecraft inspection problem is formulated as a RL problem, where the RL agent can use thrusters to change its position in space. This agent, also referred to as the deputy, is tasked with inspecting a spherical chief, which is assumed to remain stationary in Hill's frame \cite{hillsRF}. The chief is represented by a finite set of points that populate the surface of the sphere in a uniform manner, and successful completion of the task occurs once all of the points on the spherical chief have been inspected. This paper considers the illumination of the individual points on the spherical chief by a moving Sun. The details of the inspection criterion will be highlighted within this section. 

\begin{figure} [H] 
    \centering
    \includegraphics[angle=0,origin=c,width=0.55\textwidth]{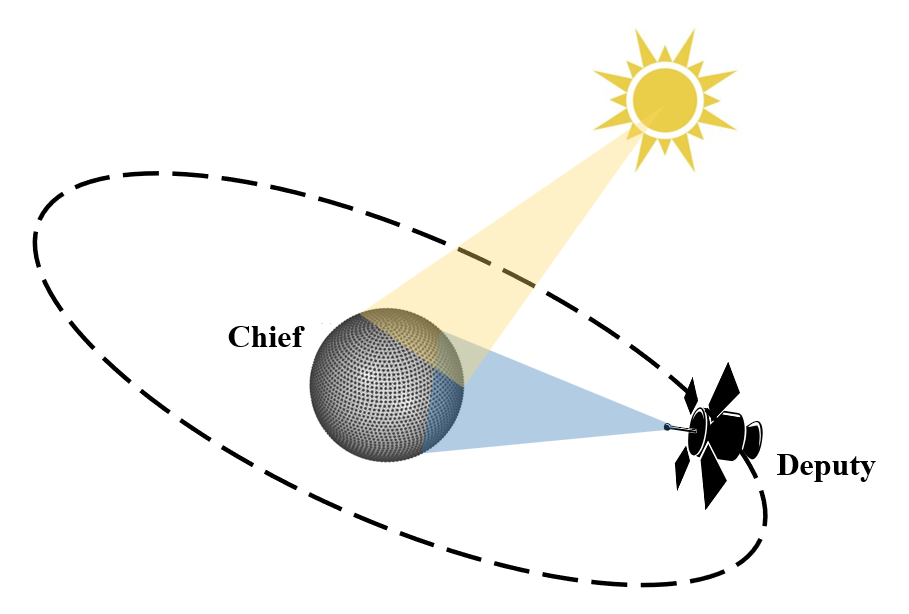}
    \caption{Inspection problem formulation.}
    \label{fig:inspection}
\end{figure}

\subsection{Dynamics}
\subsubsection{Hill's Reference Frame}
\begin{figure}[H]
    \centering
    % \centerline{\includesvg[inkscapelatex=false,width=0.5\columnwidth]{figs/cwh.svg}}
    \includegraphics[angle=0,origin=c,width=0.5\textwidth]{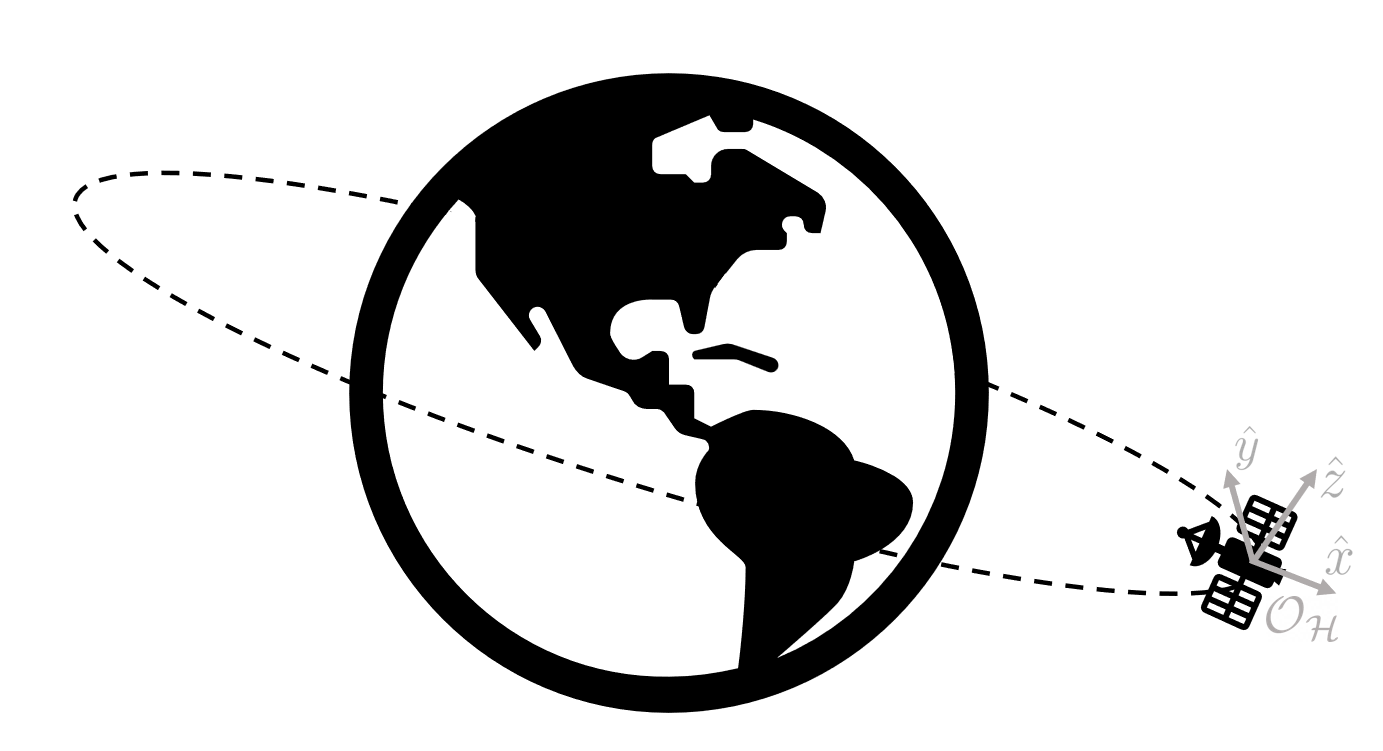}
    \caption{Hills reference frame.}
    \label{fig:hills}
\end{figure}

The dynamics and mathematical formulations are developed in Hill's reference frame shown in Fig.~\ref{fig:hills}. In this frame, the origin $\mathcal{O_H}$, is located at the center of the passive chief spacecraft, the unit vector $\hat{x}$ points away from the center of the Earth, $\hat{y}$ points in the direction of motion of the chief, and $\hat{z}$ completes the right hand rule with $\hat{x}$ and $\hat{y}$.

\subsubsection{Clohessy-Wiltshire Dynamics}
The relative motion between the deputy and chief are linearized Clohessy-Wiltshire equations \cite{CW1960}, given by, 
\begin{equation} \label{eq:CW}
    \dot{\boldsymbol{x}} = A {\boldsymbol{x}} + B\boldsymbol{u},
\end{equation}
where the state $\boldsymbol{x}=[x,y,z,\dot{x},\dot{y},\dot{z}]^T \in \mathcal{X} \subseteq \mathbb{R}^6$, the control (same as actions) $\boldsymbol{u}= [F_x,F_y,F_z]^T \in \mathcal{U} \subseteq [-u_{\rm max},u_{\rm max}]^3$, and
\begin{align}
\centering
    A = 
\begin{bmatrix} 
0 & 0 & 0 & 1 & 0 & 0 \\
0 & 0 & 0 & 0 & 1 & 0 \\
0 & 0 & 0 & 0 & 0 & 1 \\
3n^2 & 0 & 0 & 0 & 2n & 0 \\
0 & 0 & 0 & -2n & 0 & 0 \\
0 & 0 & -n^2 & 0 & 0 & 0 \\
\end{bmatrix}, 
    B = 
\begin{bmatrix} 
 0 & 0 & 0 \\
 0 & 0 & 0 \\
 0 & 0 & 0 \\
\frac{1}{m} & 0 & 0 \\
0 & \frac{1}{m} & 0 \\
0 & 0 & \frac{1}{m} \\
\end{bmatrix}.
\end{align}
Here, $m$ is the mass of the deputy and $n$ is the mean motion of the chief's orbit. In this simulation, $n$ = 0.001027 radians per second, and $m$ = 12 kilograms. Since this model doesn't account for the attitude of the deputy, it is assumed the sensor is always pointed towards the chief. The maximum thrust magnitude, $u_{\rm max}$, in this work is assumed to be 1 Newton. 

\subsubsection{Sun Dynamics} \label{sec:sundynamics}
It is assumed that the Sun is the only light source in the scene, and that the Sun stays in the $\hat{x}-\hat{y}$ plane in Hill's reference frame. The unit vector pointing from the center of the chief spacecraft to the Sun, $\hat{r}_{S}$, is defined as,
\begin{equation}
    \hat{r}_{S} = [\cos{\theta_S}, \sin{\theta_S}, 0],
\end{equation}
where $\dot{\theta}_S=-n$. It also assumed that nothing is present in the scene to obstruct the Sun rays from reaching the chief spacecraft.

\subsection{Inspection Formulation}
As mentioned above, the spherical chief is discretized into a user-specified number of points covering the surface. This is done using a tweaked version of the algorithm developed by Deserno \cite{deserno_2004}. Note that this method does not guarantee the number of points specified by the user, and the modification to the algorithm removes the dependence on the radius of the sphere. Deserno assumes a unit radius, while this assumption cannot be maintained in this application. 

\subsection{Perception Cone}

A perception cone is calculated to determine the points that are viewable by the deputy, creating a subset of all the points on the chief that could possibly be inspected at a given time. Given the position of the agent, $\boldsymbol{p}_a$, and a point on the sphere, $\boldsymbol{p}_c$, the point is an inspectable point (i.e. is in view) if and only if,
\begin{equation} \label{eq:inspectablePts}
   \frac{\boldsymbol{p}_a}{||\boldsymbol{p}_a||} \cdot \boldsymbol{p}_c \ge r_c\bigg[1 - \frac{||\boldsymbol{p}_a|| - r_c}{||\boldsymbol{p}_a||}\bigg].
\end{equation}

\noindent Here, $r_c$ is the radius of the chief. It is important to recall that for the above formulation to hold, it must be assumed that the sensor is always pointing towards the center of the chief. Only if the point is in view, will the point be evaluated using the illumination criteria outlined in the following subsections.

\subsection{Binary Ray Tracing}

In order to incentivize inspection of illuminated points, a technique derived from backward ray tracing is used. In backward ray tracing, the rays emanate from the sensor, and enter the scene, which in this case includes the chief spacecraft and the Sun. Upon a ray intersection with the chief spacecraft, a check is done to determine if this intersection point is illuminated or not. If the inspected point is both within the perception cone described above, and the point is illuminated in \textit{any capacity}, the point will be marked as inspected. For this reason, throughout this manuscript, the binary ray illumination model is also referred to as the simplified illumination model.

The illumination check can be computed quite inexpensively due to the spherical chief assumption. Starting with the point on the sphere, the line between the point and the current position of the Sun is computed. Then, the following algorithm determines if this line intersects the spherical chief. 
\\
The points on the spherical chief are defined by, 
\begin{equation}\label{e:sphere}
\| \textbf{x} - \textbf{c} \|^{2} = r_{c}^{2},
\end{equation}
with c as the center of the sphere, and $r_c$ as the radius. Next, the equation for a ray originating from the origin, \textbf{o}, spanning a distance $d$ in a direction defined by  \textbf{q} is defined with, 
\begin{equation}\label{e:rayeq}
 \textbf{x} = \textbf{o} + d\textbf{q}.
\end{equation}

\noindent Combining Eqs. \ref{e:sphere}-\ref{e:rayeq},
\begin{align}
    \| \textbf{o} + d\textbf{q} - \textbf{c} \|^{2} &= r_{c}^{2}, \\
    (\textbf{o} + d\textbf{q} - \textbf{c}) \cdot (\textbf{o} + d\textbf{q} - \textbf{c}) &= r_{c}^{2}, \\
    d^2(\textbf{q} \cdot \textbf{q}) + 2d[\textbf{q} \cdot (\textbf{o}-\textbf{c})] + (\textbf{o}-\textbf{c}) \cdot (\textbf{o}-\textbf{c}) - r_{c}^{2} &= 0.
\end{align}
Recognizing this as the quadratic equation with respect to variable $d$, $d$ is solved for as follows,

\begin{equation}
    d = \frac{-2[\textbf{q} \cdot (\textbf{o}-\textbf{c})] \pm \sqrt{(2[\textbf{q} \cdot (\textbf{o}-\textbf{c})]^2 - 4\|\textbf{q}\|^2(\| \textbf{o}-\textbf{c}\|^2 -r_{c}^{2})}}{2\|\textbf{q}\|^2}.
\end{equation}
\\
If the radicand is less than 0, $d$ will be imaginary, and thus there is no intersection. If the radicand is zero, there is only one possible value for $d$, meaning that the line is tangent to the sphere. The tangent case is thrown out, though it seldom occurs. If the radicand is greater than zero, there are two possible solutions, and if both solutions are positive, the smaller $d$ will produce the first, or closest intersection with the sphere, which is the information of interest. Thus it can be seen that the binary ray calculation can be computed extremely quickly since often only the radicand needs to be computed.

\subsection{Blinn–Phong Model}
\begin{figure}[H]
    \centering
    % \centerline{\includesvg[inkscapelatex=false,width=0.3\columnwidth]{figs/vectors_illum.svg}}
    \includegraphics[angle=0,origin=c,width=0.37\textwidth]{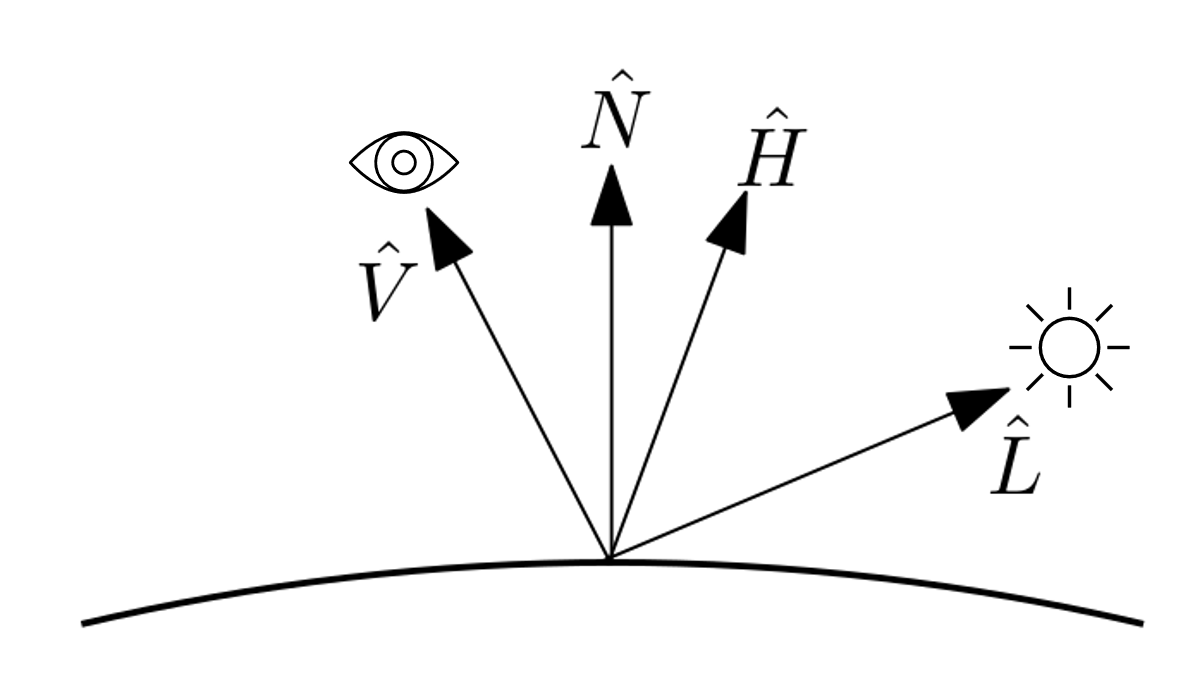}
    \caption{Vectors in Blinn-Phong illumination model.}
    \label{fig:vectors_BP}
\end{figure}

An extra layer of complexity can be added to increase the realism of the problem using a spectral illumination model often used in computer graphics. The Blinn-Phong model uses the surface properties of the chief, along with the properties of the scene light sources, and other relevant vectors to generate accurate images of the scene \cite{Phong1975IlluminationFC}. For this particular case, full scenes are not generated, the model is instead applied to individual points on the spherical chief. The first step of this calculation is the same as above: each candidate point is checked for the possibility of illumination. If the line from the point to the Sun is not obstructed, then this point is illuminated in some capacity. The second step is to use the Blinn-Phong model to determine the level of illumination. Using Eq.~\ref{eq:blinn-phong}, $I_p$ is output as a 1 x 3 array, representing the RGB illumination of that point. $\hat{N}$ is the normal to the surface, $\hat{L}$ is the vector to the light source, $\hat{V}$ is the vector to the viewer (i.e. the sensor) and the halfway vector is $\hat{H}$ where, 
\begin{equation}
    \hat{H} = \frac{\hat{L} + \hat{V}}{\| \hat{L} + \hat{V} \|}.
\end{equation}
These vectors are clearly visualized in Fig.~\ref{fig:vectors_BP}. Then,
\begin{equation} \label{eq:blinn-phong}
    I_p = k_a i_a + \sum_{l=1}^{l_{max}} (k_d(\hat{L}_l \cdot \hat{N})i_{l,d} + k_s(\hat{N} \cdot \hat{H})^{\alpha}i_{l,s}),
\end{equation}
\noindent where $k_a$, $k_d$ and $k_s$ represent the ambient, diffuse and specular reflection constants of the chief respectively. Similarly, $i_a$ represents the ambient lighting, while $i_d$ and $i_s$ are the diffuse and specular components of the light source respectively. $\alpha$ is a shininess constant for the chief.
In the case of our experiments, $l_{max}$ = 1 as the only light source is the Sun, and the surface reflection properties are given in Table~\ref{tab:surf_props}. The table also contains the intensity components of the light source, reported as normalized RGB values.

\begin{table} [H]
\centering
\caption{Chief spacecraft surface properties and light properties.}
\label{tab:surf_props}
\begin{tabular}{l|l|l|l|l|}
\cline{2-5}
                                     & \textbf{Ambient} & \textbf{Diffuse} & \textbf{Specular} & \textbf{Shininess} \\ \hline
\multicolumn{1}{|l|}{\textbf{Light}} & {[}1, 1, 1{]}    & {[}1, 1, 1{]}    & {[}1, 1, 1{]}     & N/A                \\ \hline
\multicolumn{1}{|l|}{\textbf{Chief}} & {[}.4, .4, .4{]}   & {[}.1, .1, .1{]}   & {[}1, 1, 1{]}     & 100                \\ \hline
\end{tabular}
\end{table}

\subsubsection{Threshold Criterion}

\begin{figure} [H]
    \centering
    \subfloat[\centering Sample image generated by Blinn-Phong model.]{{\includegraphics[width=5cm]{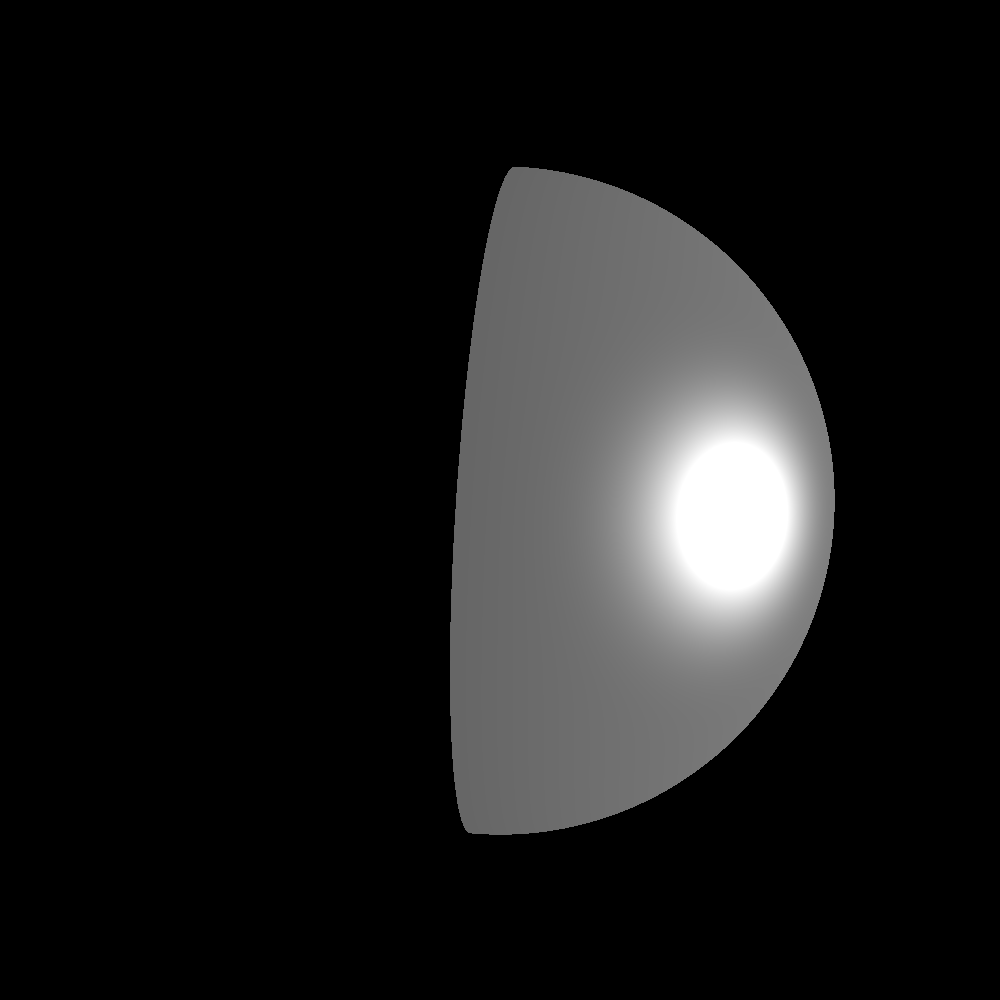} }}%
    \qquad
    \subfloat[\centering Uninspectable regions within sample image.]{{\includegraphics[width=5cm]{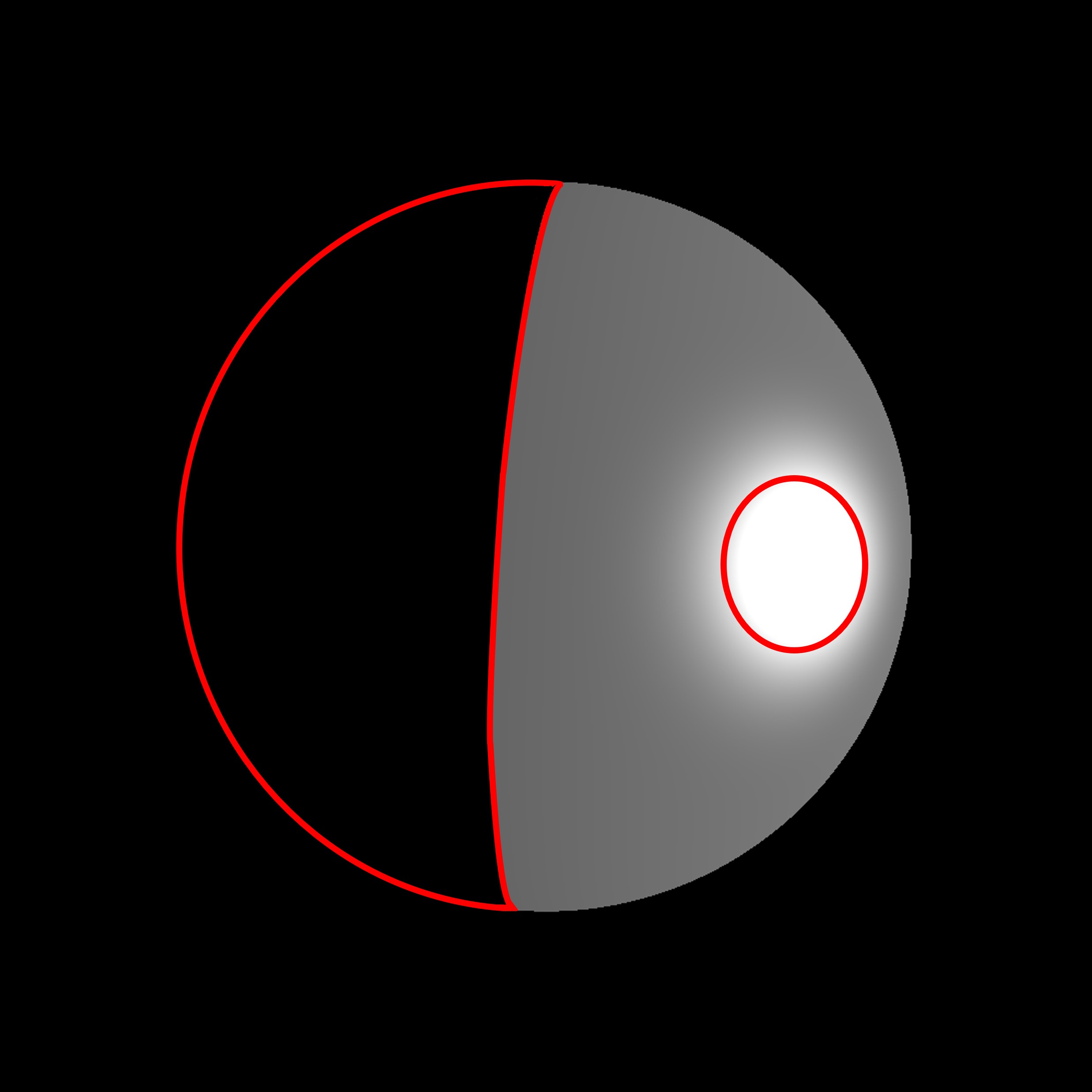} }}%
    \caption{Sample image generated by Blinn-Phong model demonstrating uninspectable regions.}%
    \label{fig:renders}%
\end{figure}

Once the RGB illumination is computed using the Blinn-Phong model, an ad hoc threshold criterion is used to determine the quality of the collected illumination data. Excessively bright or dark illumination data are considered not useful data, and are thus considered not inspected. The color of the chief is a shiny gray color intended to replicate the physical appearance of space-grade metal panels. In Fig.~\ref{fig:renders} the regions within the red outlines are too light or too dark, and are thus uninspectable. Any region with elements whose normalized RGB values exceed 0.83 are considered too bright, while regions with elements whose normalized RGB values are less than 0.2 are considered too dark.

\section{Experimental Setup}

In this application, the environment is a partially-observable environment, meaning the agent knows a sub-set of the environment states, and maps observations to actions. The observation space, $\Omega$, contains the position and velocity of the deputy spacecraft (i.e. the agent), the total number of points inspected, the Sun angle, and a direction to the largest cluster of uninspected points (Eq.~\ref{obs}). The position of the agent is encapsulated with the $x$, $y$, and $z$ position in Hill's reference frame. Similarly, the velocity of the agent is represented by linear velocities $\dot{x}$, $\dot{y}$, $\dot{z}$, in Hill's frame. The angle of the Sun relative to the $\hat{x}$-axis in Hill's frame is given by $\theta_S$. The number of points inspected ($P_{i}$) is a scalar value ranging from 0 to the maximum number of points on the chief spacecraft. Finally, the agent is also given information on the location of uninspected points, via k-means clustering.
The observation space also contains the components of the unit vector pointing to the largest cluster of uninspected points, denoted with $P_{cx}$, $P_{cy}$, $P_{cz}$.

The agent has six independent thrusters, two in opposite directions in each principal axis, and thus can take actions using those six thrusters (Eq.~\ref{actions}). In the action space, this is simplified to just three actions, where the sign of these thrust actions can be positive or negative, with magnitudes in the range [0, 1] Newtons. Additionally, the agent selects new actions and receives new observations every \textit{timestep}, which in this study is every 10 seconds. 
\begin{equation} \label{obs}
\Omega = \{x,y,z,\dot{x},\dot{y},\dot{z},\theta_S,P_{i},P_{cx},P_{cy},P_{cz}\}.
\end{equation}
\begin{equation} \label{actions}
\mathcal{A} = \{F_{x},F_{y},F_{z}\}.
\end{equation}

\subsection{Reward Function}

The actions made by the agent are evaluated at each timestep, using a scalar reward function, which is the sum of a set of distinct rewards given by Eq.~\ref{reward_sum}. The individual terms are described in detail below.
\begin{equation} \label{reward_sum}
    r = r_{points} + r_{\Delta V} + r_{crash}.
\end{equation}

\subsubsection{Observed Point Reward}

To incentivize the agent to inspect new points, $r_{points}$ is structured as follows:
\begin{equation}
  r_{points} = 0.1*(\text{number new inspected points}).
\end{equation}

\noindent The agent receives a reward of 0.1 for each new point observed.

\subsubsection{Fuel Consumption Reward}

Whilst a reinforcement learning solution does not guarantee optimal fuel performance, work can be done to mitigate excessive fuel usage and disincentivize the agent from wasteful maneuvers. At each timestep, the agent receives a negative reward based on the $\Delta V$ used, which is proportional to current fuel consumption, where,
\begin{equation}
    r_{\Delta V} = \Delta V * -w \\
\end{equation}
\begin{equation}
        \text{with~~} \Delta V = \frac{|F_{x}| + |F_{y}| + |F_{z}|}{m} \Delta t
\end{equation}
\noindent The tuning of the scalar multiplier, $w$, was a challenging task, and required much experimentation. Too high of a scalar multiplier, and the agent would learn to end the episode as quickly as possible to minimize negative cumulative reward accrued throughout each episode. Too small of a scalar multiplier, and the agent would practically ignore the $\Delta V$ penalty, opting to complete the task without considerations for minimizing thruster usage. Eventually, the $\Delta V$ penalty was implemented using a method inspired by curriculum learning. Curriculum learning is a machine learning (ML) technique whereby the ML model is trained initially using simpler scenarios and is progressively trained on more and more difficult scenarios \cite{Bengio2009}. For the context of this work, this translates to starting off with a very small $\Delta V$ penalty and gradually increasing the penalty based on performance. At the beginning of training, $w$ = 0.001, and if the mean percentage of points inspected during training exceeds 90\% for 1500 steps (approximately one training iteration), the $\Delta V$ penalty is increased by 0.00005. If the mean percentage of points inspected during training drops below 80\%, the $\Delta V$ penalty is scaled back by the same amount. Note that the minimum $w$ is 0.001. During evaluation, $w$ = 0.1, which is also the maximum value that $w$ can be. In this manner, the agent is able to initially focus on the inspection task, and gradually learns to use fuel sparingly.

\subsubsection{Crash Reward}

Clearly, crashing into the chief should be avoided. For this reason, the agent receives a negative reward if the distance from the chief is less than 10 meters plus some buffer distance, $b$, where $b$ = 5 in this study. 
\begin{equation}
     r_{crash} = 
    \begin{cases}
      -1.0 & {\sqrt{x^2+y^2+z^2} < (10 + b)} ~\text{meters}\\
      0.0 & \text{otherwise}\\
    \end{cases}  
\end{equation}

\subsection{Environment and Structure}

This research makes use of the Core Reinforcement Learning library (CoRL) developed and maintained by the Air Force Research Laboratory’s (AFRL) Autonomy Capability Team (ACT3) \cite{merrick2023corl}. This library ``is intended to enable scalable deep reinforcement learning experimentation in a manner extensible to new simulations and new ways for the learning agents to interact with them'' and makes use of RLlib \cite{liang2018rllib}. The reinforcement learning agents are trained using a NN with an input layer of size 11 and an output layer of size 3 (i.e. continuous observations to actions). There are two hidden layers, and 256 hidden nodes with hyperbolic tangent (tanh) as the activation function. The RL agent takes actions in the simulated environment and receives a scalar reward based on the actions taken. Over the course of training, the agent learns behavior that will maximize reward over many environment interactions (i.e. timesteps). The discount factor, $\gamma$, is a scalar value influencing the importance of future rewards and is 0.99 throughout training. 

These environment interactions are divided into \textit{episodes}. A new episode begins when the environment is reset. During a reset, initial conditions for the agent and the environment are randomly selected. The agent is spawned some distance between 50 and 100 meters from the chief with an azimuth angle from 0 to 2$\pi$ radians and an elevation angle between -$\frac{\pi}{2}$ and $\frac{\pi}{2}$ radians. These polar coordinates are then converted into Cartesian coordinates. Similarly, the direction of the agent's initial velocity is obtained using polar coordinates with an azimuth angle from 0 to 2$\pi$ radians and an elevation angle between -$\frac{\pi}{2}$ and $\frac{\pi}{2}$ radians. The magnitude of this initial velocity is between 0 and 0.3 meters per second.

The Sun is also initialized with a relative angle $\theta_S$ between 0 and 2$\pi$ radians. There are four events that will trigger the end of an episode:
1) the time horizon is reached (i.e. maximum number of timesteps), 2) the agent crashes into the chief, 3) the agent's distance from the chief exceeds 800 meters, and 4) the agent inspects all points (task completion). The maximum number of timesteps for this study is 1224, which is the time required for two Sun orbits. Two full Sun orbits gives the agent opportunity to inspect all the points on the chief even if it is initialized in unfavorable conditions (i.e. on the opposite side of the chief as the Sun).

\section{Simulation Results}

This section presents the performance of the RL agents throughout the training process as well as the performance of the final models and representative trajectories upon deployment of the final policies.

\subsection{Training Performance}

Due to the stochastic nature of RL, 10 agents were trained using the same reward function but with different random seeds. The learned policies are periodically evaluated during training and the interquartile mean (IQM) is used to analyze agent performance. The IQM has been shown to be a better metric for summarizing training performance of RL agents \cite{agarwal2022deep}. The IQM is robust to outlier scores but more statistically efficient than the median.

\subsubsection{Binary Ray Illumination}

\begin{figure}[htp]
\centering
\centering
\includegraphics[angle=0,origin=c,width=0.48\textwidth]{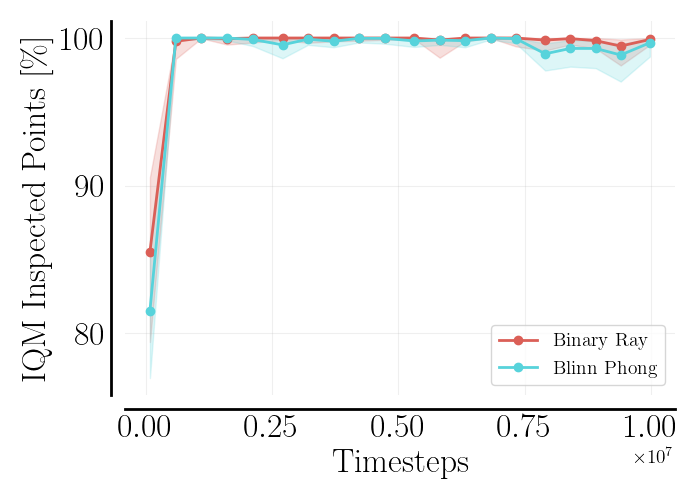}\quad
\includegraphics[angle=0,origin=c,width=0.48\textwidth]{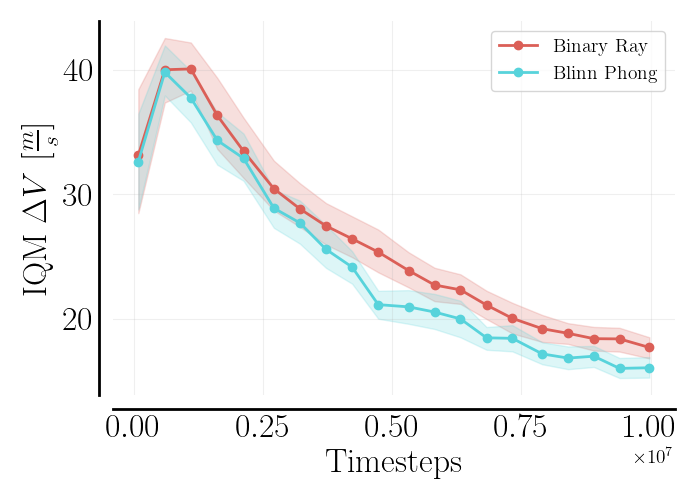}
\bigskip
\includegraphics[angle=0,origin=c,width=0.48\textwidth]{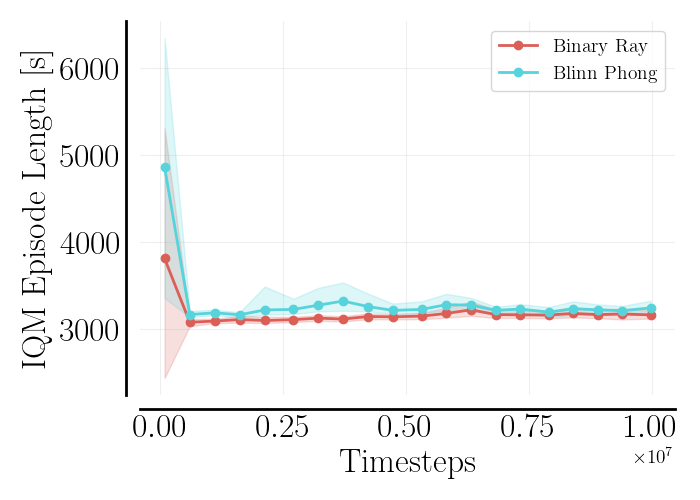}\quad
\includegraphics[angle=0,origin=c,width=0.48\textwidth]{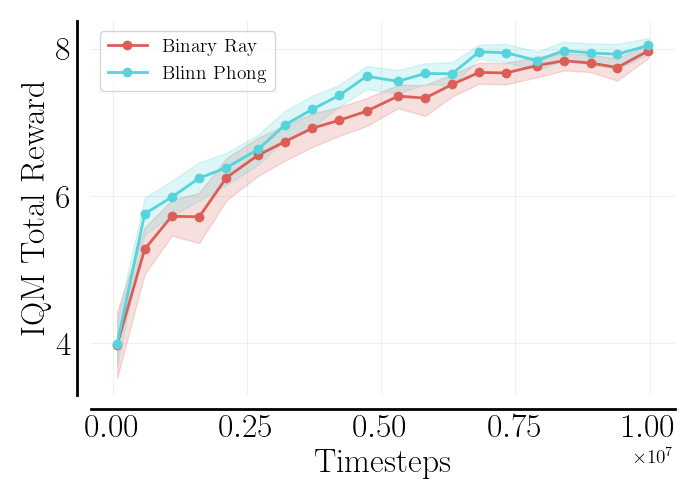}

\caption{Performance metrics using simplified illumination model and Blinn-Phong illumination model evaluated throughout training using an IQM over 10 seeds.}
\label{fig:evalTraining_both}
\end{figure}

Fig.~\ref{fig:evalTraining_both} shows the interquartile mean (IQM) percentage of points inspected, $\Delta V$ used to complete the task, episode length, and total reward received, all as a function of total timesteps. The 95\% confidence interval for the IQM is also plotted on the figures as the shaded regions. Each seed was trained for 10 million timesteps, and these four metrics were evaluated throughout training by deploying the current policy in the environment for 10 episodes and recording the statistics. Fig.~\ref{fig:evalTraining_both} plots the metrics for both the simplified illumination model and the spectral model.

At the beginning of training, the percentage of inspected points is low while the agent explores the environment. As training timesteps increase, percentage of points inspected increase, along with the $\Delta V$ utilized by the agent. The $\Delta V$ consumption quickly reaches a maximum, and steadily declines, with the exception of a few minor jumps. This is because in tandem, the percentage of inspected points increases and thus the magnitude of the penalty associated with $\Delta V$ consumption increases due to the structure of the reward scheduling scheme. The increased cost associated with $\Delta V$ consumption incentivises the agent to conserve fuel. 

\subsubsection{Blinn-Phong Illumination}

In Fig.~\ref{fig:evalTraining_both} the same four evaluation metrics are plotted evaluated throughout the training process for the spectral illumination model. Again, the 95\% confidence interval for the IQM is also plotted on the figures as the shaded regions. As expected, the RL agent initially does not inspect a large percentage of points. Additionally, the fuel usage is relatively high, and the IQM of episode length is also high, while total reward accrued remains low. The RL agent quickly learns to perform the task well, inspecting nearly all the points within the first 250,000 timesteps. However, the fuel required to adequetly perform the inspection task is very high. This is because the reward scheduling procedure associated with $\Delta V$ consumption starts off with a relatively low cost for fuel. As the training process progresses, the IQM for inspected points remains near maximum, while the fuel used gradually decreases. The IQM of episode length quickly flattens out to around 3200 seconds. Notably, the minimum amount of time the task can be completed in is approximately 3060 seconds, or the time required for the Sun to complete half a full revolution. Lastly, the total reward seems to flatten out, which is a good indicator that the policy has matured.

In general the training performance of the agents using each illumination model is very similar, both following very similar trends. Initially, both have a very large spread in the IQM of episode length. Both sets of agents also converge to very similar IQM's of total accumulated reward. However, the binary ray illumination model has a slightly better performance for percentage of inspected points, and uses slightly more $\Delta V$ than the Blinn-Phong illumination model. 

\subsection{Final Model Performance}

After the training has terminated, an RL model is then ``frozen" meaning that the internal weights are no longer modified, and the deterministic policy can be evaluated by deploying it in the environment. On deployment, the initial conditions (agent initial position, velocity and Sun angle) are randomized to obtain a fair representation of the space. Each trained final policy was evaluated 100 times, and the results are tabularized below for both illumination models.

\subsubsection{Binary Ray Illumination}

Table~\ref{tab:eval_binray} summarizes the performance of the final models over the 10 seeds using the simplified illumination model. The same four metrics of percentage of points inspected, $\Delta V$, episode length and total reward are used. The tabularized results are representative of what the trained RL agent would do under random initial conditions in this environment, and are the representation of the best model obtained. The percentage of points inspected is very high, at 99.83\%, with a $\Delta V$ usage that is much lower than the initial policies. Again, the episode length is very close to the minimum amount of time required to inspect all the points on the spherical chief. 

\begin{table}[H]
\centering
\caption{Final policy evaluation performance over 100 trials for 10 random seeds using simplified illumination model.}
\centering

\begin{tabular}{|l|c|c|}
\hline
\multicolumn{1}{|c|}{\textbf{Metric}} & \textbf{IQM} & \textbf{\begin{tabular}[c]{@{}c@{}}95\%\\  Confidence Interval\end{tabular}} \\ \hline
Inspected Points {[}\%{]}             & 99.83        & {[}99.74, 99.91{]}                                                           \\ \hline
$\Delta V$ {[}m/s{]}                     & 18.08        & {[}17.80, 18.37{]}                                                           \\ \hline
Episode Length {[}s{]}                & 3217        & {[}3199, 3236{]}                                                           \\ \hline
Total Reward                          & 7.885        & {[}7.856, 7.913{]}                                                           \\ \hline
\end{tabular}

\label{tab:eval_binray} 
\end{table}

\begin{figure}[H]
    \centering
    % \centerline{\includesvg[inkscapelatex=false,width=1\columnwidth]{figs/model_performance/binray/seed_8827_episode_plot.svg}}
    \includegraphics[angle=0,origin=c,width=1\textwidth]{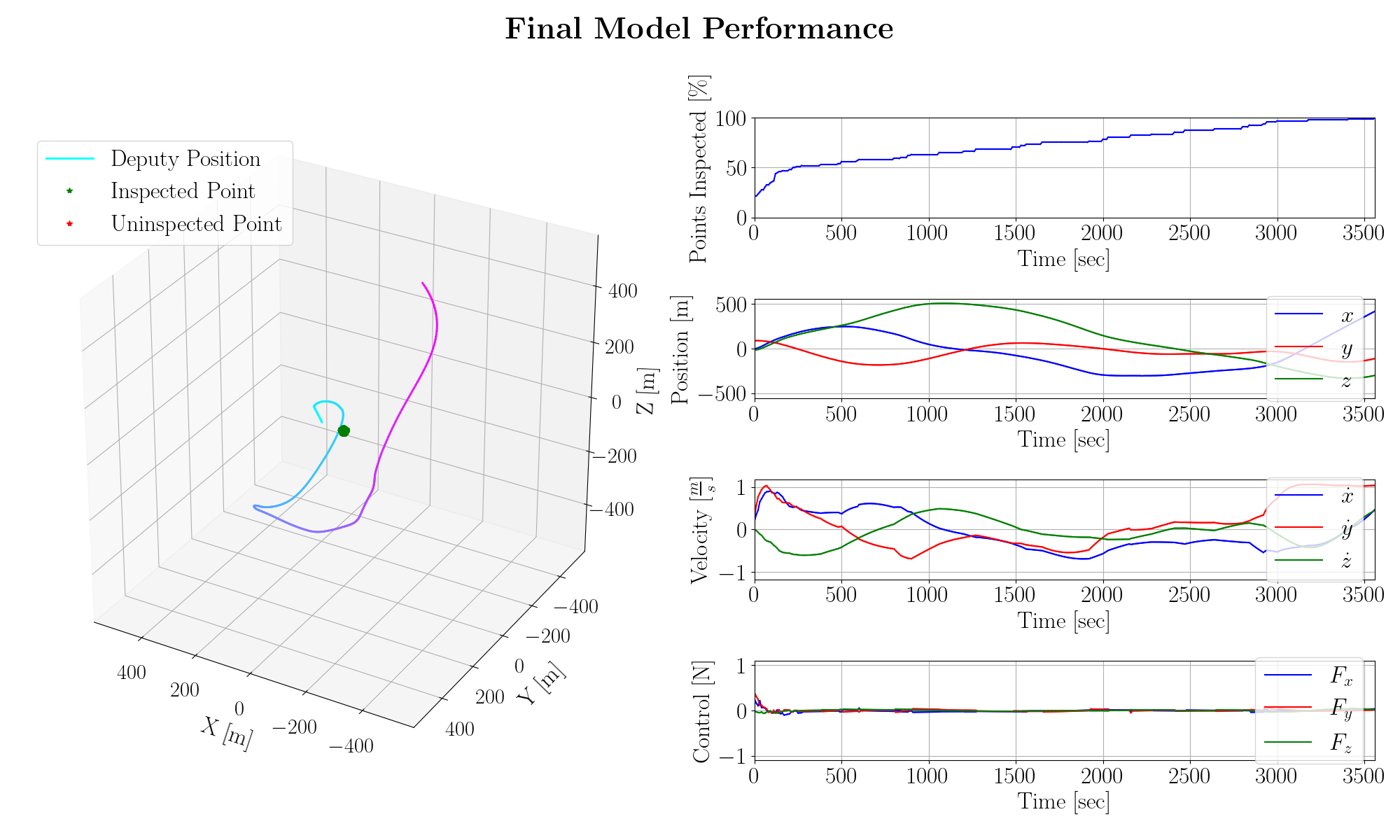}
    \caption{Representative final model performance using simplified illumination model.}
    \label{fig:results1_binray}
\end{figure}

\begin{figure}[H]
    \centering
    % \centerline{\includesvg[inkscapelatex=false,width=1\columnwidth]{figs/model_performance/binray/seed_739_episode_plot.svg}}
    \includegraphics[angle=0,origin=c,width=1\textwidth]{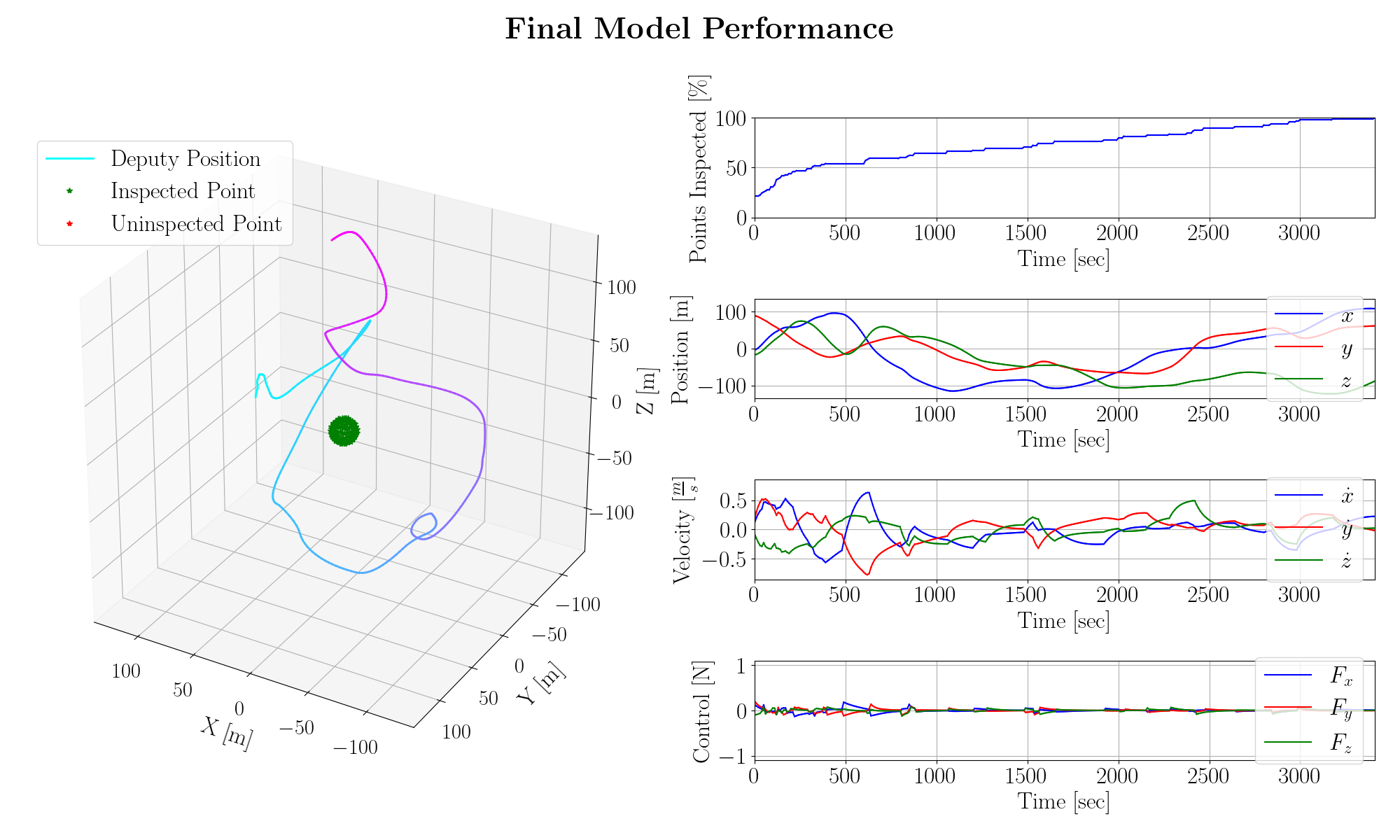}
    \caption{Representative final model performance using simplified illumination model.}
    \label{fig:results2_binray}
\end{figure}

A visualization framework showing agent actions, observations and trajectory was built to better understand the solution method converged upon by the final policies. This visualization framework creates both static plots and animations. Figs.~\ref{fig:results1_binray} and~\ref{fig:results2_binray} plot the percentage of points inspected, the components of position and velocity of the agent, and the actions at each timestep. The plots also feature the trajectory of the agent in space on the left hand side. The color gradient of the trajectory represents time elapsed, with cyan being the beginning of the episode. In both representative examples, the agent successfully inspected all points, while having low $\Delta V$ consumption. One can see that the agent uses its thrusters sparsely, making use of the natural motion, and gradually circles the chief as the Sun rotates, gaining more and more new information as the episode progresses. Animations capturing the motion of the free-flying RL agent as the Sun rotates are available as well\footnote{\href{https://youtu.be/-OpmZ7y24zQ}{https://youtu.be/-OpmZ7y24zQ.}}. These animations more clearly demonstrate the agent's motion in 3D space. 

\subsubsection{Blinn-Phong Illumination}

As above, the same analysis was performed for the final models using the spectral illumination model. The IQM for the same four metrics are very similar to the simplified model and are summarized in Table~\ref{tab:eval_blinnphong}. Notably, the agent inspection performance is slightly worse than for the case of the simplified model. This is somewhat anticipated since all things equal, the spectral illumination model should produce a set of inspectable points that are a subset of the points produced by the simplified illumination model. However, the IQM for $\Delta V$ utilized is slightly lower than for the agents trained with the simplified illumination model. This could also mean that the agents learned to value fuel slightly more than point inspection.

\begin{table}[H]
\centering
\caption{Final policy evaluation performance over 100 trials for 10 random seeds using Blinn-Phong illumination model.}
\begin{tabular}{|l|c|c|}
\hline
\multicolumn{1}{|c|}{\textbf{Metric}} & \textbf{IQM} & \textbf{\begin{tabular}[c]{@{}c@{}}95\%\\  Confidence Interval\end{tabular}} \\ \hline
Inspected Points {[}\%{]}             & 98.82        & {[}98.45, 99.13{]}                \\ \hline
$\Delta V$ {[}m/s{]}                     & 16.25        & {[}16.01, 16.50{]}                \\ \hline
Episode Length {[}s{]}                & 3181        & {[}3159, 3202{]}                \\ \hline
Total Reward                          & 7.890        & {[}7.856, 7.921{]}                \\ \hline
\end{tabular}  \label{tab:eval_blinnphong} 
\end{table}

\begin{figure}[H]
    \centering
    \includegraphics[angle=0,origin=c,width=1\textwidth]{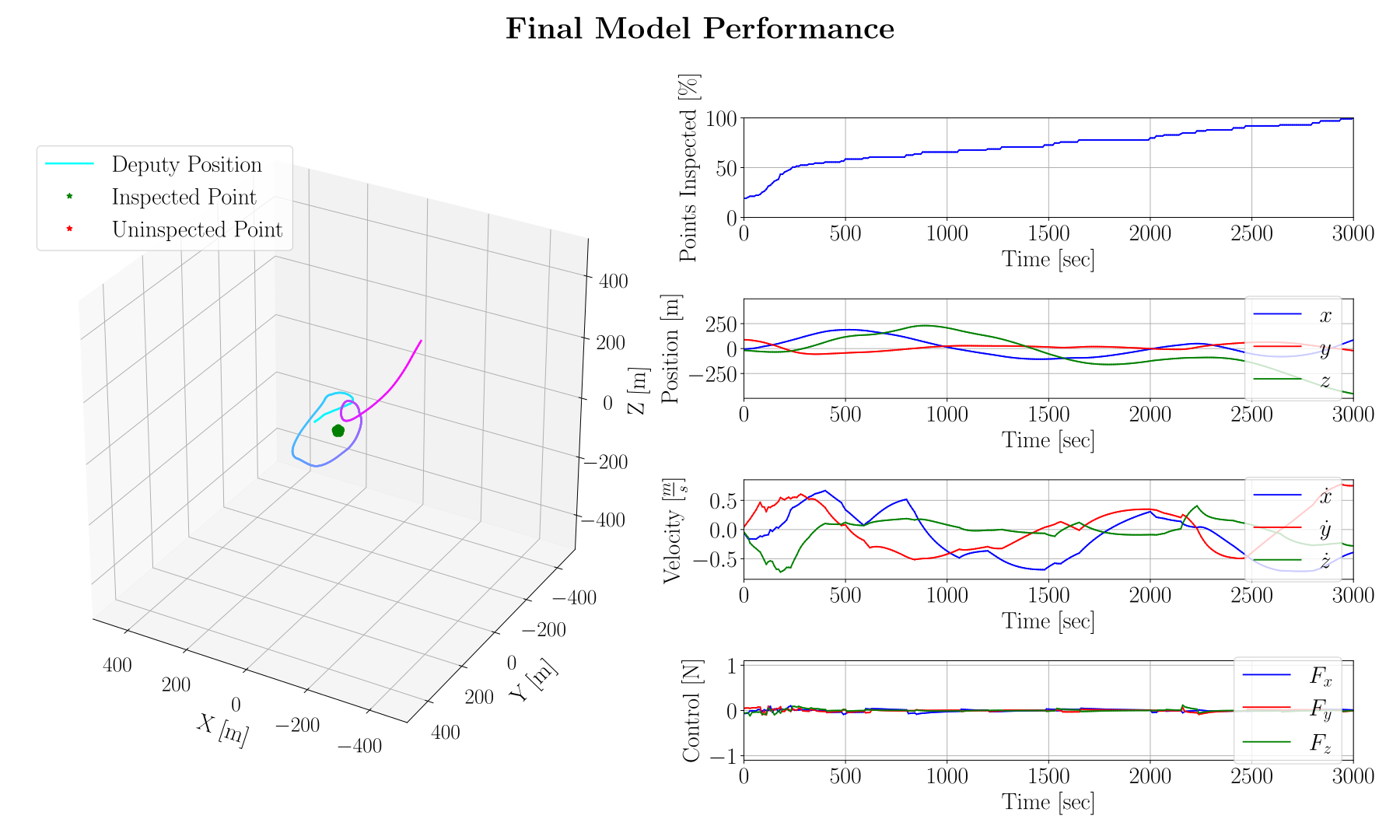}
    % \centerline{\includesvg[inkscapelatex=false,width=1\columnwidth]{figs/model_performance/blinnphong/bp_seed_8827_episode_plot.svg}}
    \caption{Representative final model performance using Blinn-Phong illumination model.}
    \label{fig:results1_blinnphong}
\end{figure}

\begin{figure}[H]
    \centering
    \includegraphics[angle=0,origin=c,width=1\textwidth]{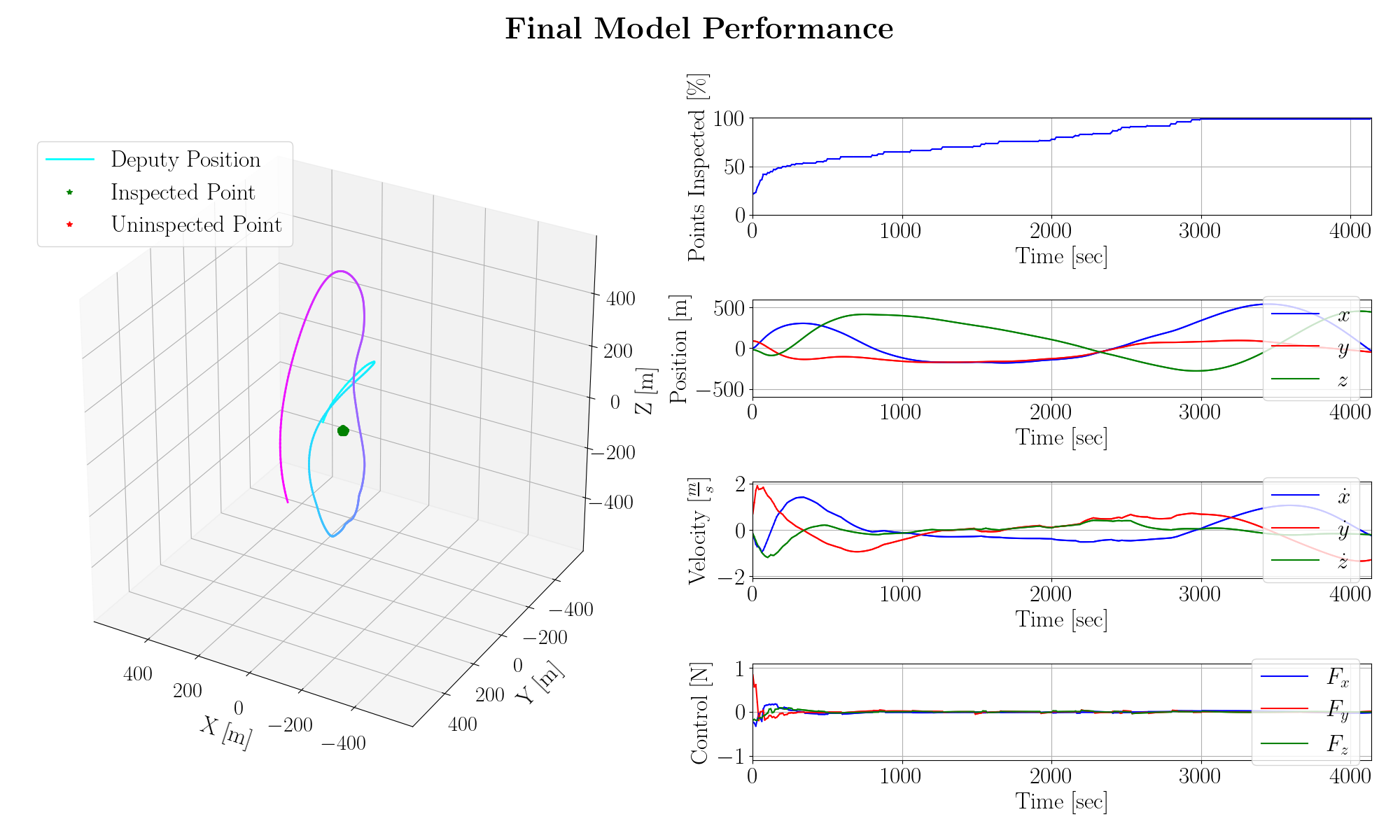}
    % \centerline{\includesvg[inkscapelatex=false,width=1\columnwidth]{figs/model_performance/blinnphong/seed_9110_episode_plot.svg}}
    \caption{Representative final model performance using Blinn-Phong illumination model.}
    \label{fig:results2_blinnphong}
\end{figure}

Figs.~\ref{fig:results1_blinnphong},~\ref{fig:results2_blinnphong} plot the percent points inspected, the components of position and velocity of the agent, and the actions at each timestep. In both examples, all the points are inspected in near minimal time, while having low $\Delta V$ consumption, using the thrusters sparsely, again making use of the natural motion. In Fig.~\ref{fig:results2_blinnphong}, the agent inspects nearly all the points around 3000 seconds into the simulation, and is able to position itself as the Sun rotates such that it finally inspects the remaining points on the spacecraft with minimal additional control usage. Once it inspects all the points the episode terminates.

\section{Conclusion}

This paper presented a solution to the single-agent spacecraft inspection problem with illumination considerations. A Proximal Policy Optimization deep reinforcement learning controller was developed for a 3 DOF spacecraft obeying the linear dynamics dictated by the Clohessy-Wiltshire equations, with thruster capability in each principal direction. Based upon the results presented in the paper, the following conclusions are drawn: 
\begin{enumerate}
    \itemsep-.07em 
    \item The proposed illumination model provides a sufficient formulation for quick and efficient illumination computation in the process of training a reinforcement learning policy for satellite inspection. 
    \item The reinforcement learning agent presented here is an effective method to solve the inspection problem with the proposed illumination framework. Results in a simulated environment using the spectral illumination model demonstrated the robustness of the learned policy, showing over 10 random seeds, RL agents were able to successfully inspect an IQM of 98.82\% of points on the chief over 100 trials using only low-level actions.
\end{enumerate}

This paper made some important assumptions that will be relaxed in future work. Firstly, the occlusion from other bodies, including the Moon and Earth, will be considered. Secondly, the authors will explore the possibility of a non-spherical chief and a 6 DOF agent, with full attitude information and control authority. The possibility of a tumbling chief spacecraft may also be an interesting research topic to address.

\section*{Acknowledgment}

The views expressed are those of the authors and do not reflect the official guidance or position of the United States Government, the Department of Defense or of the United States Air Force.

\bibliographystyle{AAS_publication}   % Number the references.
\bibliography{references}

\end{document}